\documentclass[a4]{article}
  \usepackage{cite}

\usepackage{amsmath}
\usepackage{amsfonts}
\usepackage{bbm}
\usepackage{hyperref}

\usepackage{algorithm, algorithmic}

\usepackage{subfigure}

\usepackage{stfloats}
\usepackage{todonotes}

\usepackage{color}

\usepackage{acronym}
\acrodef{MDP}{Markov Decision Process}
\acrodefplural{MDP}[MDPs]{Markov Decision Processes}
\acrodef{PCA}{Principal Component Analysis}
\acrodef{RL}{Reinforcement learning}
\acrodef{GLS}{Global Longitudinal Strain}
\acrodef{MCTS}{Montecarlo Tree Search}
\acrodef{DAG}{directed acyclic graph}
\acrodef{AUC}{area under curve}
\acrodef{UCI}{University of California at Irving}
\acrodef{REPS}{Relative Entropy Policy Search}

\usepackage{authblk}
\usepackage{orcidlink}
\title{Sampling-guided exploration of active feature selection policies}

\author[1,2]{Gabriel Bernardino\orcidlink{0000-0001-8741-2566} \thanks{Corresponding Author: gabriel.bernardino@upf.edu}}
\author[1]{Anders Jonsson \orcidlink{0000-0002-5756-7847}}
\author[2]{Patrick Clarysse \orcidlink{0000-0002-5495-7655}}
\author[2,3]{Nicolas Duchateau \orcidlink{0000-0001-8803-2004}} 

\affil[1]{Department of Engineering, Universitat Pompeu Fabra, Barcelona, Spain}
\affil[2]{Univ Lyon, Universit\'e Claude Bernard Lyon 1, INSA-Lyon, CNRS, Inserm, CREATIS UMR 5220, U1294, F-69621, Lyon, France}
\affil[3]{Institut Universitaire de France (IUF)}

\begin{document}
\maketitle

\begin{abstract}
Determining the most appropriate features for machine learning predictive models is challenging regarding performance and feature acquisition costs. 
In particular, global feature choice is limited given that some features will only benefit a subset of instances. In previous work, we proposed a reinforcement learning approach to sequentially recommend which modality to acquire next to reach the best information/cost ratio, based on the instance-specific information already acquired. We formulated the problem as a Markov Decision Process where the state's dimensionality changes during the episode, avoiding data imputation, contrary to 
existing works. 
However, this only allowed processing a small number of features, as all possible combinations of features were considered.
Here, we address these   limitations with two contributions: 1) we expand our framework to larger datasets with a heuristic-based strategy that focuses on the most promising feature combinations, and  2) we introduce a post-fit regularisation strategy that reduces the number of different feature combinations, leading to compact sequences of decisions. We tested our method on four binary classification datasets (one involving high-dimensional variables), the largest of which had 56 features and 4500 samples. We obtained better performance than  state-of-the-art methods, both in terms of accuracy and policy complexity. 
\end{abstract}



\section{Introduction}\label{sec:introduction}
Computer-aided diagnosis has potential to improve clinical workflows, allowing to take advantage of large amounts of data to detect patterns associated with disease. As in human-based diagnosis, it involves the combination of different sources of information, such as tests or imaging modalities, and, therefore, algorithms have been proposed to map all the information to a common space\cite{Li2019ALearning,Antelmi2019SparseData,Sanchez-Martinez2017CharacterizationLearning}. A drawback of 
most approaches is that they generally require homogeneous and complete datasets where all descriptors are acquired,
while in a real setting not all available modalities contribute for diagnosis of all individuals. 
Besides, the acquisition of unnecessary modalities should be minimized, since they are associated with costs \cite{Joynt2015HealthMedicine,Huang2015OveruseServices,Albarqouni2023OveruseReview}
Such costs can involve the economic price of acquiring the data itself, but can also include other concepts: for instance in medical diagnosis, many tests are dangerous or uncomfortable for the patient such as tissue biopsy or nuclear imaging. Despite the importance of minimizing acquisition cost at inference time, most machine learning methods still consider input data as a fixed part of the problem. 


Classical methods for feature selection, such as lasso regularization which promotes coefficient sparsity, work at a population level, selecting a subset of features that are enough to do the prediction for every data instance \cite{Meier2008TheRegression,Pudjihartono2022APrediction}. However, this approach is not optimal since some features are relevant for only part of the instances, and thus feature selection needs to be performed for each instance individually at inference time to obtain a cost-effective data acquisition. 

On the other hand, other approaches also address 
this by 
sequentially proposing instance-specific features to be acquired, similar to a clinician referring to an informative test, minimizing the cost-information gain trade-off \cite{Dulac-Arnold2011Datum-wiseSparsity, Min2014FeatureConstraint, Ma2018EDDI:VAE,Yin2020ReinforcementAcquisition}. A challenge for these approaches is that acquired features evolve during the process; thus estimators have to cope with inputs of varying dimensionality. 
A possible solution is to treat not-acquired data as missing, and use data imputation strategies to derive a complete view from partial information; however, these solutions might produce bias and fail when features cannot be estimated from each other.
Some authors pre-specified different models, each associated with a different budget for feature acquisition, \cite{Erion2022AApplications,Bernardino2022HierarchicalRemodelling,Bolukbasi2016ResourcePrediction}, and used some basic features to decide the optimal model. This approach obtained good results in practical applications, but requires domain knowledge and assumes that the cost is the main factor to select features, while there can be pathologies presenting several etiologies that require different features for their diagnosis.
Another approach is to estimate the joint probability distribution of all features, as done in \cite{Li2020ActiveModels}, but this is a difficult task that requires high computational power and large datasets. Other approaches, such as \cite{NIPS2015_c0a271bc}, try to build different models for each possible combination of features, but the number of possible features combinations grows exponentially, and the approach turns computationally infeasible for medium/large number of features.

Several works addressed active feature selection using greedy strategies, iteratively acquiring the feature that maximises the expected information gain in a stepwise fashion \cite{Ma2018EDDI:VAE,Covert2023LearningSelection}.
However, these might avoid acquiring features that are not informative by themselves, but become very informative when combined with other variables. To derive optimal acquisition planning, which considers not only immediate gain but also future improvements, the acquisition sequence can be seen as an \ac{MDP}. An \ac{MDP} is a mathematical setting to represent sequential processes, called episodes. An episode is a sequence of states and rewards, where each reward and state depends on the previous state and an action that was taken at this step. Reinforcement learning aims to estimate optimal policies, which given a state compute the best action to take for maximising final rewards. 
However, reinforcement learning algorithms typically require a large amount of training instances, while for many clinical problems only a few hundreds of instances are available. 

\begin{figure*}
  \includegraphics[width = \textwidth]{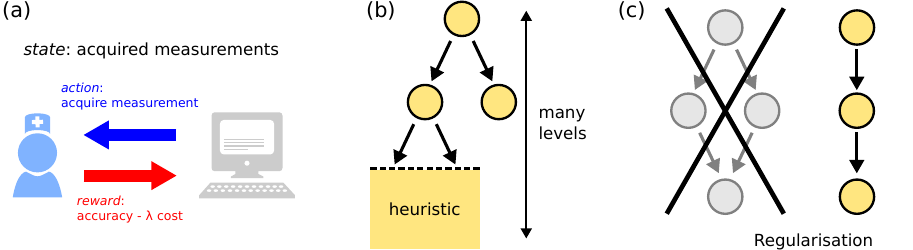}
  \caption{Overview of our proposed reinforcement learning approach for active feature selection, in the context of medical data. (a) The decision of using a new acquisition/measurement represents the action at each state, guided by a reward that combines the diagnosis accuracy and the cost associated to acquiring these features, as introduced in our previous work \cite{Bernardino2022ReinforcementDiagnosis} and consolidated here (Sections~\ref{sec:MDP} and \ref{sec:activeFeatureSelection}. 
  (b) Proposed policy exploration based on a heuristic, very relevant for large number of features (Section~\ref{sec:policyExploration}).
  (c) Proposed regularisation strategy to reduce the total number of different action sequences (Section~\ref{sec:policyRegularisation}).}
  \label{fig:Overview}
\end{figure*}

\subsubsection*{Related works}

In previous work \cite{Bernardino2022ReinforcementDiagnosis}, we demonstrated the relevance of \ac{RL} for active feature selection on specific medical imaging problems involving a small number of high-dimensional variables and costs associated to the input features, but the approach required considering all possible combinations. Here, we substantially extend this work to datasets involving a medium number of variables, addressing key issues related to this increase in the amount of variables, and compare to the state-of-the-art methods discussed below.

EDDI \cite{Ma2018EDDI:VAE} is an algorithm based on data imputation using variational autoencoders, which are deep learning models that can estimate missing features from partial views. 
This is done via a mask that sets to 0 the values of the non-acquired data.
This model can also be used for prediction by treating the objective label as a variable that is always unobserved during inference. To obtain the next feature to acquire, different instances with possible values of each candidate feature are sampled, and the one that reduces the most uncertainty over the target variable is chosen.

The clustering heuristic proposed by \emph{Wang et al.} \cite{NIPS2015_c0a271bc} shares many similarities with our approach. 
The authors used an offline heuristic to group a subset of the features in sets, limiting the amount of effective features during dynamic selection, while our approach restricts the search space. Also, their approach groups different features, forcing them to be acquired simultaneously, thus reducing the effective number of features. This feature grouping is the same for the whole population and is learnt offline before inference, by forcing that each instance can be correctly classified by at least one of these groups. The choice of groups is also constrained by the features' cost, whose sum cannot exceed a prescribed maximum. This forces that the total amount of different features acquired among the different individuals is bounded. A problem is that this heuristic does not promote the features to be split in different groups, and it tends to concentrate all selected features in few groups, reducing the capacity of the method to select features at an instance level.

\subsubsection*{Contributions}

An overview of our methodology, and the experiments, is given in Figure~\ref{fig:Overview}. More specifically, our main contributions are: 
\begin{itemize}
    \item A graph-exploration heuristic that focuses on the most promising feature combinations, and incrementally builds a policy constrained to the visited spaces (Section~\ref{sec:policyExploration}),

    \item A regularisation strategy that reduces the total number of different action sequences, improving the model interpretability and reducing overfitting (Section~\ref{sec:policyRegularisation}),
    \item Extensive evaluation of the accuracy and policy cost on three public datasets and a private dataset, three of them involving medical data (Section~\ref{sec:experiments}). 
\end{itemize}

\section{Methodology}

\subsection{Markov Decision Process}
\label{sec:MDP}

In this subsection, we provide a quick summary of the general notation of \ac{RL}
we use in this paper.
An \ac{MDP} \cite{Sutton2018ReinforcementIntroduction} is a tuple $(\mathcal{S}, T, R, \mathcal{A})$, where:
\begin{itemize}
\item $\mathcal{S}$ and $\mathcal{A}$ represent the set of possible states and actions respectively,
\item $T$ is the transition function, which associates each combination of the current state and chosen action to a probability distribution over the next state,
\item $R$ is the reward function, which associates the current state, chosen action and next state to a probability of the reward obtained. 
\end{itemize}

\noindent An \emph{episode} is a trajectory of states, actions, and rewards:
\begin{equation}
(s_0, a_0, r_0, s_1, a_1, r_1, \dots , s_{N-1}, a_{N-1}, r_{N - 1}, s_N, r_N), \nonumber
\end{equation}
where the sequences of states and rewards respectively follow $T$ and  $R$, and $s_N$ is an ending state\footnote{In a more general case, episodes might not have an ending state and might continue  indefinitely,  however in our case, we will only consider finite episodes.}. 

The objective is to find an optimal policy $\pi: \mathcal{S} \to \mathcal{A}$, which is a mapping from the state space to the action space that maximises the \emph{return} (sum of rewards during the episode $ret = \sum_{i=0}^N r_i$).
This optimal policy can be found by computing the optimal $Q$-functions, which contain the expected return achievable for trajectories starting at state $s$ and whose next action is $a$. Optimal $Q$-functions have to satisfy Bellman's optimality equation:
\begin{equation}
\label{eq:Qsa}
    Q(s, a) = \underset{s'}{\mathbb{E}} \left[ R(s,a, s') + \max_{a'} Q(s', a') \right],
\end{equation}
where $s'$ corresponds to the next state. Note that this recurrence formulation (since $Q$ appears both in the left and right sides of Equation~\ref{eq:Qsa}) is challenging to solve, and typically iterative sampling methods need to be used.

Finally, a policy $\pi$ that maximises the $Q$-value is optimal:
\begin{equation}
    \pi(s) = \underset{a}{\mathrm{argmax}}\hspace{3pt} Q(s,a).
\end{equation}

\subsection{Active feature selection as a Markov Decision Process}
\label{sec:activeFeatureSelection}

In this subsection we  define active feature selection as an \ac{MDP}. Each episode corresponds to the diagnosis of a specific patient (instance), and thus the transitions and reward are deterministic once the patient is fixed.
At each step, the possible actions to choose are to either acquire a specific new feature (for example, a given measurement from a specific imaging modality), or to finish the exploration and diagnose. The state space encodes all information acquired so far. Note that the transition and reward functions during inference are stochastic, since they depend on the current state (which only contains the acquired features), and there might be patients who exactly coincide in the acquired features but have different numerical values for the non-acquired features; thus adding uncertainty to the label/missing information prediction from a partial state.

The methods exposed in this subsection correspond to a more detailed explanation of our previous work \cite{Bernardino2022ReinforcementDiagnosis}.
In the next two subsections, we detail the extensions to this framework that consist of the main contributions of the current paper.

\subsubsection{Superstates}

Unlike the most common frameworks used by other authors, we do not work in a fused connected space that represents all features together, but in a collection of disconnected domains, one for each possible combination of acquired features, avoiding the need for data imputation. We will call each of these domains a \emph{superstate}. This means our state space will consist of elements with distinct dimensionality. Note that transitions between superstates are completely deterministic and do not depend on the instance, just on the action taken (since they only encode which modalities were acquired, but not the measurement observed). Superstates (and states) form a \ac{DAG} denoted $G$, with every superstate as a node, and edges between nodes if and only if there is a direct transition between them. Formally, the state space $\mathcal{S}$ is defined as:
\begin{equation}
    \mathcal{S} = \underset{f \subseteq \mathcal{F}} {\dot\bigcup} S_f,
\end{equation}
where $f$ is a subset of features taken from $\mathcal{F}$ (set of all possible features), $\dot\bigcup$ is the disjoint union, and $S_f$ encodes all possible combinations of values taken by the features $f$. Namely, $S_f \sim \mathbb{R}^{N_{dim(f)}}$, where $N_{dim(f)}$ is the sum of all feature dimensionalities ($N_{dim(f)} = |f|$ if all features are scalars). 

At the population level and in particular for the regression used during the policy training, we consider the matrix $\mathbf{S}_f \in \mathbb{R}^{N_{patients} \times N_{dim(f)}}$, which groups the data for the $N_{patients}$ instances for a given subset of features $f$.

\subsubsection{Rewards}

In our case, the reward function is defined for each instance (patient) as:
\begin{align}
        R(s_i, a_i, s_i+1) = 
    \begin{cases} 
      0 & \text{if}~i \neq N_{end}, \\
      acc_i - \lambda ~ cost_i & \text{otherwise}
   \end{cases}   \label{eq:reward}
\end{align}

where $acc_i$ is a random variable denoting if the instance was correctly classified by an external classifier with features available at state $s_i$ (the accuracy), $cost_i$ is the cost associated to acquiring these features, and $\lambda$ is a weighting factor. This formulation is possible as our episodes are guaranteed to finish after $| \mathcal{F} |$ steps.

At the population level, $\mathbf{acc}^f \in \mathbb{R}^{N_{patients}}$ is a vector grouping the accuracy for all instances, and $cost^f = cost_i$ as the cost depends on the features in $f$ and not instances.

\subsubsection{Solving Bellman's equation}

Since we can sort the states, given the \ac{DAG} structure,
a direct solution of Bellman's optimality equation can be obtained without sampling as explained in our previous work \cite{Bernardino2022ReinforcementDiagnosis}. Algorithm \ref{alg:cap:train} summarizes this procedure. Superstates are visited in postorder. The expectation in the right hand side of Equation~\ref{eq:Qsa} is taken over the whole population, and the optimal $Q$-values are estimated using a regression model, as $s$ belongs to a continuous state space $\mathcal{S}$. Since the state space includes elements of different dimensions, the algorithm trains a $Q$-estimator for each superstate (combination of features) $f$ and action $a$, each trained using the features $\mathbf{S}_f$ (the matrix containing the features available at said superstate for the whole population). These estimators are trained independently. We denote them $Q_S$, the subindex corresponding to the superstate $S_f$. Once the optimal $Q$-estimators are constructed, we derive the optimal policy by taking the action with the highest $Q$-value. 

\subsubsection{Values}

We can also define the value of each state, measuring the future rewards accessible from this state, as:
\begin{equation}
    V(s) = \max_a Q(s,a).
\end{equation}
For actions corresponding to acquiring a feature that has been already acquired, the associated $Q$-value is set to $-\infty$.

At the population level, values are grouped into a vector $\mathbf{V}_f \in \mathbb{R}^{N_{patients}}$, which takes the values of $\max_a Q_f(\mathbf{S}_f,a)$.

\begin{algorithm}
    \caption{Pseudo-code for training the policy. It requires two external functions, \emph{train\_regressor} and \emph{predict}, which train and evaluate an arbitrary regression model, respectively. Bold variables refer to population data consisting of $N_{patients}$ instances: $\mathbf{S}_f$ is the matrix with all features corresponding to the superstate $S_f$, 
    $\mathbf{V}_f$ is the vector of corresponding values, and $\mathbf{acc}^f$ and $cost^f$ are the population accuracy and cost. $\max_a$ refers to the instance-wise maximum, and $FINISH$ stands for the finishing action.} \label{alg:cap:train}
    
    \begin{algorithmic}[1]
      \FORALL {$f \subseteq \mathcal{F}$, sorted by $|f|$} 
        \STATE $Q_f(\cdot, FINISH) \gets \mathbf{acc}^f - \lambda~cost^f$
    
        \FORALL {$a | a \neq FINISH$}
            \STATE $Q_f(\cdot, a) \gets \infty$
            \IF {$a \notin f$}
                \STATE $Q_f(\cdot, a) \gets train\_regressor(\mathbf{S}_f, \mathbf{V}_{f \cup \{a\}})$
            \ENDIF
        \ENDFOR
        \STATE $\mathbf{V}_f \gets \max_a Q_{f}.predict(\mathbf{S}_f, a)$
      \ENDFOR
    \end{algorithmic} 
\end{algorithm}

\subsubsection{Additional considerations}

To add more stability and avoid overfitting, we can split the training set in two parts, one used to estimate the optimal $Q$-values and the other to fit the $V$-estimators. This is analogous to double $Q$-learning. However, we have not observed any benefit  for this strategy in the experiments performed on our data.

\subsection{Sampling-guided policy exploration for large number of features}
\label{sec:policyExploration}

A drawback of the previously defined \ac{MDP} is that optimising a policy involves training exponentially-many (on the number of modalities) estimators, since a $Q$-model, and potentially a classifier, is learnt for each  combination of modalities 
(referred to as superstate). 
However, many superstates correspond to suboptimal feature combinations that can be removed without affecting the performance of the optimal policy. In this subsection, we present a sampling-based method to identify the most promising superstates, and build a restricted policy that only considers a subset of superstates; thus avoiding the training of low-quality $Q$-value estimators.  

To select a subset of superstates $\mathcal{T}$ capturing the highest value regions, we use a root-to-leafs approach. This set $\mathcal{T}$ is initialised with the root of the superstate graph $\mathcal{G}$, and iteratively expanded, adding one direct descendant of $\mathcal{T}$ at a time. At each step, we compute an optimal policy $\pi_\mathcal{T}$, which only visits the superstates of $\mathcal{T}$. The decision of which descendant to add is guided by a sampling estimation. A possible choice is to estimate the expected return of episodes starting at the candidate superstate $S_f$, noted as $ \widehat{Ret}(S_f) $, by randomly generating a sequence of superstates $S_k$, with $k \supset f$. For each $k$ and each element $j$ of the population, we compute its value $s^j_k$ (based on the prediction accuracy and cost), with the features available at superstate $S_k$. The final reward is obtained by estimating the maximum value for each individual, and averaging over all individuals:
\begin{align}
\begin{split}
\label{eq:estimationReturn}
    \widehat{Ret}(S_f) &= \mathbb{E}_j [\max_k V(s^j_k) ] \\
    &=  \mathbb{E}_j [\max_k [acc^j_k - \lambda cost^k] ]
\end{split}
\end{align}
Algorithm \ref{alg:cap} summarizes the steps of this sequential heuristic expansion of $\mathcal{T}$. Our algorithm bears resemblance to heuristic solvers of the graph traversal problem, where the objective is to find a minimal path from a root to an objective, as 
in \cite{Hart1968APaths}. Unlike the exhaustive exploration of all nodes seen in Dijkstra's algorithm, heuristic graph traversal selectively analyzes a subset of nodes. In their approach, nodes are sequentially chosen and incorporated into the subset through a heuristic process that estimates the distance to the target from each potential node.
\begin{algorithm}
    \caption{Pseudo-code for sample-guided policy exploration, based on a heuristic-based search. $N_{max}$ is the number of maximum allowed explorations, $\mathcal{G}$ refers to the full graph of superstates.}\label{alg:cap}
    \begin{algorithmic}[1]
      \STATE $\mathcal{T} \gets \{ \emptyset \}$ \COMMENT{Initialise}
      \STATE $\mathcal{N} \gets adjacents(\emptyset )$ \COMMENT{Set with the neighbours of the current graph}
      \STATE \COMMENT{Iteratively choose the best $N_{max}$ node subset}
      \WHILE {$|\mathcal{T}| < N_{max}$ }
            \STATE \COMMENT{Compute the heuristic for all neighbouring nodes}
            \FORALL {$n \in \mathcal{N}$}
                \STATE $\widehat{Ret(S_n)} \gets heuristic(S_n)$, as in Eq \ref{eq:estimationReturn}
            \ENDFOR
            \STATE $n_{chosen} \gets \arg\max_{n \in \mathcal{N}} \}$
            \STATE $\mathcal{T} \gets \mathcal{T} \cup \{ n_{chosen} \}$
            \STATE Remove $n_{chosen}$ from $\mathcal{N}$
            \STATE $\mathcal{N} \gets \mathcal{N}  \cup \{m \in adjacents(n_{chosen}) \cap m \notin \mathcal{N} \}$
      \ENDWHILE
    \end{algorithmic} 
\end{algorithm}

\subsubsection{Sampling-based estimation of the return} \label{subsection:valueEstimation}

To compute the heuristic, our algorithm requires an estimation of the return from a given superstate $S_i$. An option is sampling: a series of superstates reachable from $S_i$ are randomly selected, and the returns of finishing at those superstates are computed. Note that this is possible since in our \ac{MDP} the return only depends on the finishing state, but not the path towards it (Equation~\ref{eq:reward}).

We take the instance-wise maximum value of the different visited superstates, assuming that the optimal policy  is able to direct each instance to the highest-reward superstate. Note that as the total number of superstates with exactly $n$ acquired features follows Newton's binomial, a naive sampling strategy over-represents states close to $N_{features}/2$. To compensate for this bias,  we use a logarithmic sampling strategy, first sampling the number of acquired features of the superstate uniformly, and afterwards the specific features.

\subsubsection{Policy update}

Each time a new superstate is added to the restricted set~$\mathcal{T}$, the new optimal policy $\pi_\mathcal{T}$ needs to be updated. Algorithm~\ref{alg:policy_opt_update} shows an efficient computation using the previous policy $\pi_{\mathcal{T}'}$. The algorithm is very similar to the usual policy optimisation, in which the predicted values for each of the actions are propagated from the leafs to the root. However, we stop the propagation if the instance-wise maximum value $V$ of a superstate has not changed at this iteration.

\begin{algorithm}
    \caption{Pseudo-code for updating restricted policy $\pi_\mathcal{T}$ after adding a new superstate $S_n$. }
    \label{alg:policy_opt_update}
    \begin{algorithmic}[1]
      \REQUIRE {New node $S_{new}$, and its end-values $V_{new}$ }
      \STATE $Update \gets Queue\{new\}$
      \WHILE {$Update \neq \emptyset$}
        \STATE Choose and remove a superstate $n$ from the $Update$ set.
        \STATE Update $\mathbf{V}_n$ as in Algorithm \ref{alg:cap:train}
        \IF{ $\mathbf{V}_n$ has not changed}
        \STATE \textbf{continue}
        \ELSE
        \FORALL {$m \in ancesters(n)$}
          \STATE $a$ is the action needed to transition from superstate $S_m$ to $S_n$
          \STATE $Q_m(\cdot, a) \gets train\_regressor(\mathbf{S}_m, \mathbf{V}_n)$
        \ENDFOR
        \ENDIF
      \ENDWHILE
    \end{algorithmic} 
\end{algorithm}

\subsection{Sparse policy regularisation}
\label{sec:policyRegularisation}

In this section, we  introduce a regularisation technique for reducing the number of different superstates visited for all the population. Compact policies that involve less superstates not only  are more interpretable, but  can also be visualised more easily, and are  less prone to overfitting. To promote sparsity, we add a new term to the policy loss  (which until now corresponds to the expected return). The section is organised as follows. First, we define a policy-level regularisation term. Then, we break it into instance-level contributions to the reward. Finally, we propose an iterative algorithm to solve the regularised \ac{MDP}. The theoretical background of these derivations can be found in Appendix~\ref{sec:AppendixA}, where we show  that it can be seen as a particular case of \ac{REPS}, as described in \cite{Peters2010}.

\subsubsection{Policy-level regularisation}

As in \ac{REPS}, we consider the occupancy measure of a given policy $\pi$, a "function" that associates each subset of states to a real number expressing the expected (over all the instances) number of visits per episode at each (set of) state following policy $\pi$. We denote this measure $\mu$ instead of $\mu_\pi$ by abusing notation.

With this notation, we can express an $L_0$ regularisation term that penalises the number of visited superstates (from a population-level) as:
\begin{equation}
    E_{reg} (\mu) = \sum_n  H(\mu(S_p)),
\end{equation}
where $H$ is the Heaviside function and $S_f \subseteq \mathcal{S}$ refers to superset associated to the features $f$. 

Since minimising a function that includes the function $H$ produces theoretical and numerical problems, as it is equal to $1$ almost everywhere, we instead optimise its smooth relaxation $L_q$, with $0 < q < 1$:

\begin{equation}
    E_{reg}^q(\mu) =  \sum |\mu(S_f)|^q.
\end{equation}

\subsubsection{Virtual reward}

To be integrated to the \ac{RL} framework, the previous regularisation term needs to be transformed from its policy-level formulation to an instance-level definition. We 
therefore add a virtual term to the original reward, which  only depends on the new superstate of the transition, representing the variation of $H$ under a new policy that takes that action. If the change to the policy is infinitesimal, this variation is proportional to the derivative of $E_{reg}^q$. So, the virtual reward $R_{virtual}$ is
defined as:
\begin{equation}
\label{eq:virtualReward}
    R_{virtual}(s,a,s') = \alpha~q~\mu(S_f')^{1 - q},
\end{equation}
with $\alpha$ an hyper-parameter to smooth the problem and make it solvable by continuous optimisation techniques. This virtual reward 
is added to the traditional reward $R$ defined in Equation~\ref{eq:virtualReward}, which is non-zero only at the final state and is based on the cost and the correctness of the prediction.

\subsubsection{Optimisation}

We initialise the algorithm with a non-regularised  policy, and then run an iterative scheme, in which we add the virtual reward previously presented, using the measure from the current iteration's policy. We can then optimise the new \ac{MDP}, and obtain a new policy, which visits less superstates. This is repeated until convergence.

Given that the regularisation term is non-concave and the optimisation is a maximisation, we cannot ensure convergence towards a global maxima (and it might not be unique), but only to a local one.  In our setting, the $\alpha$ introduced in Equation~\ref{eq:virtualReward} has a role similar to as the step-size of a gradient method, and therefore can influence the convergence rate, if convergence is achieved.

\section{Experimental setting}
\label{sec:experiments}

\subsection{Datasets}

We used three publicly available datasets from the \ac{UCI} repository to test our methodology: 
\begin{itemize}
    \item[$\bullet$] the "Heart" dataset (predicting whether a patient admitted to the intensive care unit suffers from myocardial infarction based on vital signs and laboratory biomarkers), 
    \item[$\bullet$] the "Wisconsin breast cancer" dataset (detect cancer in breast histological images from predefined biomarkers),
    \item[$\bullet$] the "spambase" dataset (classify spam mail based on the appearance frequency of different words).
\end{itemize}
From these datasets, only the "Heart" one included real costs of each feature.  

We also used a private dataset ("HTA") consisting of 5 different temporal measurements of the cardiac motion and flow along a full cycle, acquired using 2D echocardiography of the left ventricle. These measurements included the blood velocity at the aortic and mitral valves (acquired using Doppler ultrasound), tissue velocity at the basal free wall (Tissue Doppler Imaging), and speckle tracking derived deformation of the cardiac muscle, both at the global and regional levels (focusing on the septal free wall, a region that typically presents alteration in hypertension-related pathologies). For each of these measurements, we extracted its waveform which were used in the analysis. These waveforms were temporally aligned and resampled to be in correspondence among different individuals, and afterwards were treated as a multidimensional feature: we used PCA to reduce their dimensionality before the \ac{RL} method we propose, keeping the first 5 modes. 
Table \ref{table:datasets_description} summarizes the main characteristics of these datasets. For all, we used the \ac{AUC} and mean accuracy as metrics in independent test-sets, repeated in different bootstrap samples.

\begin{table*}
\centering
\caption{Characteristics of the used datasets.}
\label{table:datasets_description}
\tiny
\begin{tabular}{c|ccccccc}
Dataset & public & contains costs & \# Features  & \# Samples & \# Class prevalence [\%] & Figure \\
\hline 
Heart & yes & yes & 11 & 531 & 61 & \ref{fig:SOTA},\ref{fig:heteroCosts},\ref{fig:diff_lambdas} \\
Breast & yes & no & 8 & 683 & 34 & \ref{fig:SOTA}  \\
Spam & yes & no & 56 & 4601 & 39 & \ref{fig:SOTA},\ref{fig:differentTraining} \\
HTA & no & no & 5 (multi-dim.) & 248 & 70 & \ref{fig:SOTA},\ref{fig:exp:policies_reg},\ref{fig:reg} \\
\end{tabular}
\end{table*}

\subsection{Base estimators}

Our state-action value estimator was a kernel ridge regression with an RBF Gaussian kernel for each feature, using the standard choice of bandwidth. We used Nyström approximation to reduce the 
footprint of computing the kernel. For the final classification, which determines the predicted class during inference and the value of each instance at training, we chose a kernel logistic regression model. To improve computational performance, the weights were shared among the different superstates, but only the weights associated to acquired features were used during inference, and the non-acquired features are set to the population average.

To avoid overfitting, we used a 10 fold cross-validation approach to obtain the predicted value of each instance, and this prediction was used to compute the reward at the final nodes.

\subsection{Comparisons with SOTA}

We compared our method to the following state-of-the-art algorithms, discussed in the Introduction (Section \ref{sec:introduction}): 
\begin{itemize}
    \item[$\bullet$] The EDDI algorithm (Efficient Dynamic Discovery of high value Information) \cite{Ma2018EDDI:VAE},
    \item[$\bullet$] The clustering heuristic proposed by \emph{Wang et al.} \cite{NIPS2015_c0a271bc} 
\end{itemize}

We extended the EDDI original algorithm to work with multidimensional variables, which is the case for the HTA dataset, by allowing sampling and acquiring a group of scalar features simultaneously (corresponding to a single multidimensional-feature). We reimplemented Wang's algorithm, using the same base estimators as in our algorithm, instead the decision trees they proposed in their paper.  

Finally, since both EDDI and Wang's heuristic require a maximum budget, instead of minimising the population average acquisition cost, we reformulated our method to have a maximal number of acquisitions to allow comparisons. This was implemented by forcing the policy to take the end-action from all states where the maximum amount of modalities has already been acquired. 

\section{Results}

\subsection{Diagnosis accuracy}

\begin{figure}
  \includegraphics[width = \columnwidth]{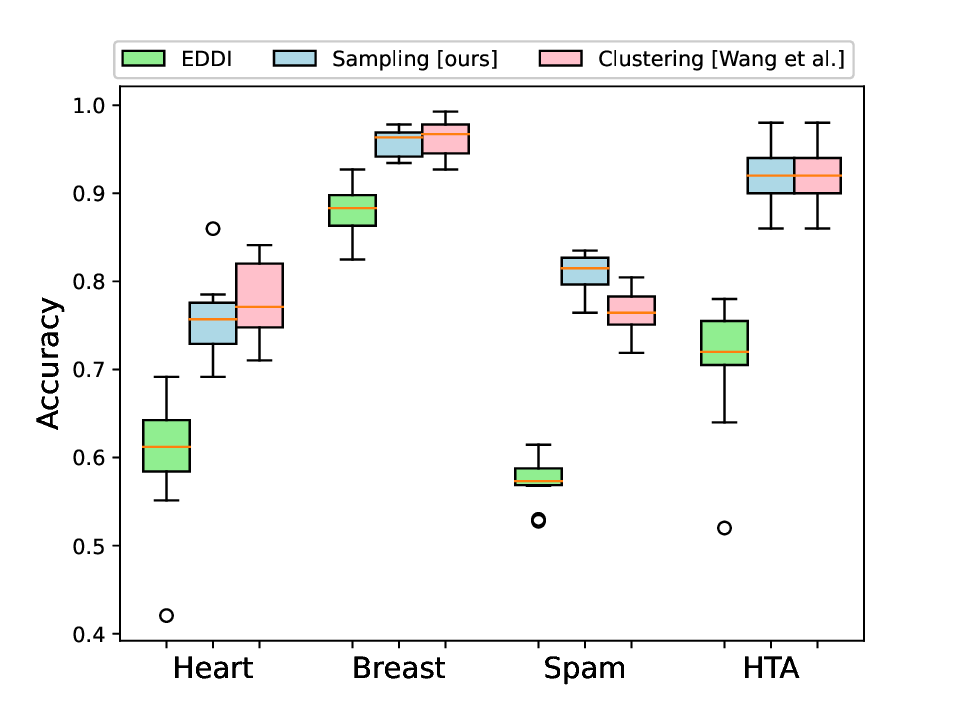}
  \caption{Diagnostic accuracy, measured by the AUC, of our method compared to EDDI and Wang's methods, on the four studied datasets (result of acquiring a maximum of 3 features).}
    \label{fig:SOTA}
\end{figure}

Figure \ref{fig:SOTA} compares the diagnostic accuracy of our method to EDDI and Wang's methods, on the four studied datasets. It corresponds to the result of acquiring a maximum of 3 features. Both our algorithm and Wang's clustering approach outperform EDDI on the three datasets with a small number (around 10) of features, namely "Heart", "Breast", and "HTA"), with a much better performance in the multidimensional one ("HTA"). EDDI fails in HTA, since HTA has high-dimensional data with highly uncorrelated features and EDDI fails to impute the  missing features from the available ones. We must highlight, however, that the available public implementation of EDDI is not optimised for predicting discrete variables, which might have resulted in a loss of prediction accuracy. 

Wang's clustering approach slightly outperforms ours on the three datasets with a small number of features, but fails on the dataset with a higher number of features ("Spam", which contains around 50 features). This inadequacy stems from the heuristic employed by Wang's approach, which imposes a maximum limit on the total number of features acquired by the policy. Consequently, this constraint proves overly restrictive for datasets with a larger number of variables, as each feature plays a critical role in accurately classifying diverse individuals. 

\subsection{Feature's acquisition cost}

In this section, we focus on the "Heart" dataset, as it was the only dataset with real costs available, and compare our method with Wang's approach in a more realistic setting, in which we try to minimise the average cost instead of having a hard limit on the number of modalities per individual. Costs represent the economic cost, in Canadian dollars, of performing each test. 
The features consist of demographic information (age and medical history), the vital signs at admission (resting ECG and blood pressure), blood parameters that require a laboratory test,
and an exercise test\footnote{The original dataset also includes imaging, but we have discarded this feature since it was only available at one of the four centers that provided data.}.

\begin{figure*}
  \includegraphics[width = \textwidth]{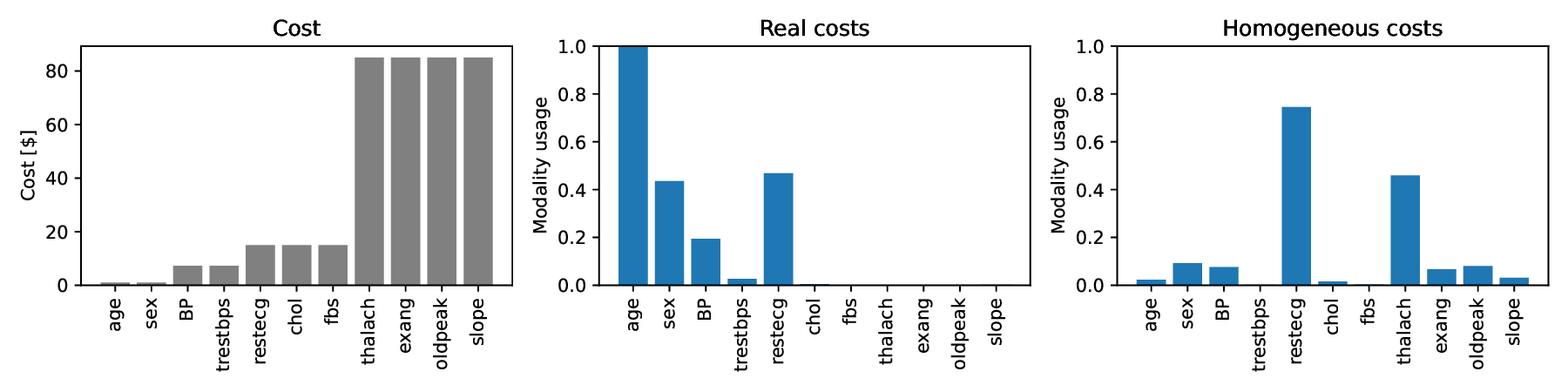}
  \caption{Modality usage (probability that a modality is acquired,averaged over the full testing population in 10 different train-test splits) in the "Heart" dataset. Features are sorted by their cost (from left to right), whose values are summarized in the first plot.  We can see that when real costs are used (middle column), the algorithm prefers cheaper variables (the ones located on the left area of the plot), while when each variable has the same cost (right column), then more expensive variables are used (the ones located on the right area of the plot).}
\label{fig:heteroCosts}
\end{figure*}

We first compared the result of setting the same cost to each feature versus using the real ones (Figure \ref{fig:heteroCosts}).
This comparison is particularly relevant since there are some features, such as "slope" that are very informative, but costly to acquire (since it involves  the patient undergoing an exercise testing). Results show that when the costs are set to homogeneous, the algorithm prefers to take expensive variables (third column), but when costs are non-homogeneous, it will prefer cheaper variables (second column).

\begin{figure}
\centering
  \includegraphics[width = 0.9\columnwidth]{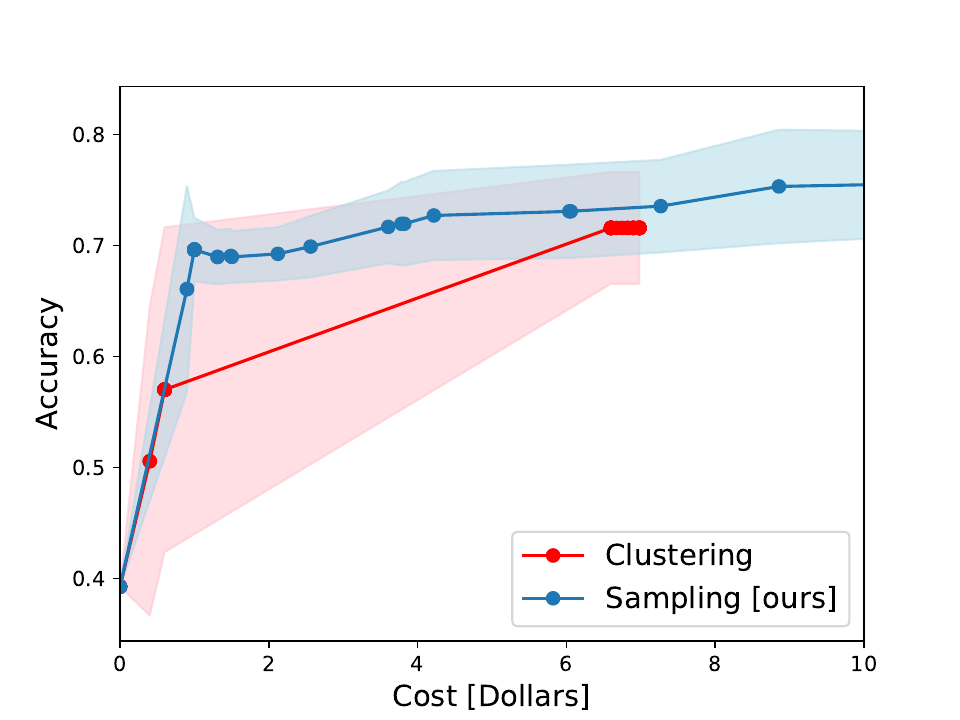}
  \caption{Mean accuracy as a function of the mean cost for our method ("sampling", blue) and Wang's ("clustering", red), for different values of the parameter $\lambda$, for the "Heart" dataset. We tested different $\lambda$ which means steering the balance between cost and accuracy in the reward (each represented as a point). Our approach allows smoother transition, where expensive features are acquired only for difficult cases, whereas Wang's approach has a more dichotomic behaviour between high and low cost. }
  \label{fig:diff_lambdas}
\end{figure}

Then, 
we compared our method with Wang's clustering approach, testing for various values of the $\lambda$ hyperparameter, 
therefore obtaining several balances of accuracy and cost (Figure \ref{fig:diff_lambdas}). A limitation of Wang's approach is that, even if the policy learning allows selectively acquiring groups of features, features tend to be concentrated in a few groups. Indeed, 
Wang's clustering does not promote distributing features in different groups. 
This leads to results similar 
to using $L_0$ feature selection at a population-level. Results show that our method outperforms Wang's, 
the accuracy being higher for similar average cost. 
A remarkable difference is that increases in cost and accuracy in our method are gradual, since only a portion of the population can access 
the expensive features; in contrast, 
Wang's method increments are more stepwise, as groups are obtained at a population level.

\subsection{Analysis of the heuristic exploration}

We studied how the heuristic allowed a faster exploration in the Spam dataset, which had the highest number of variables (56 features). Thus, we compared to a dummy strategy, in which the same number of nodes was added by randomly selecting the next node to open.

At each step, we added a new  superstate to our restricted set, neighbouring our current subgraph, based on the estimation given by the heuristic. This 
was 
repeated a prescribed number of times (200 open nodes), but a stop criterion could be added when no changes on the policy return are observed.

\begin{figure*}
  \includegraphics[width = 0.99\textwidth]{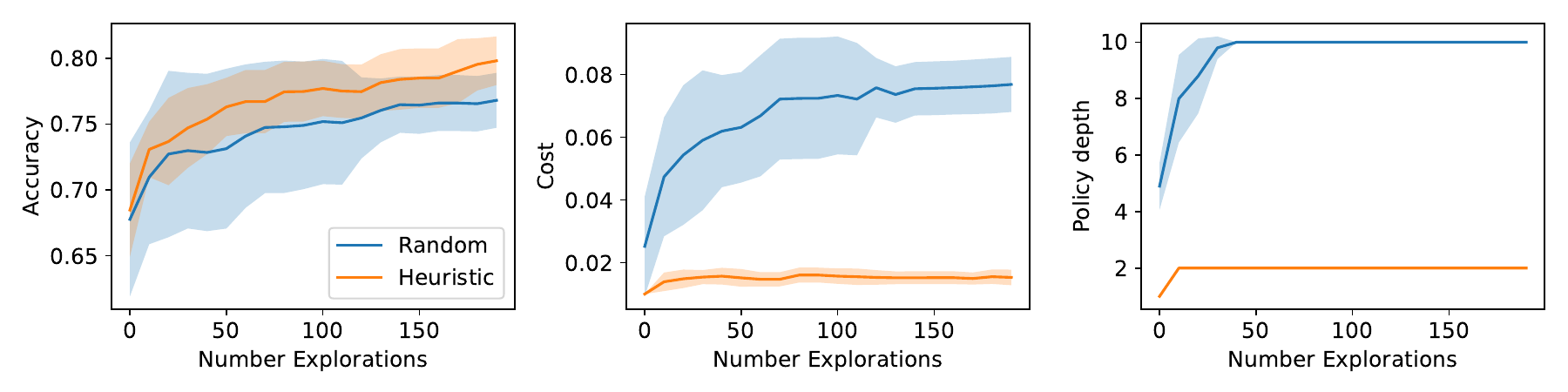}
  \includegraphics[width = 0.99\textwidth]{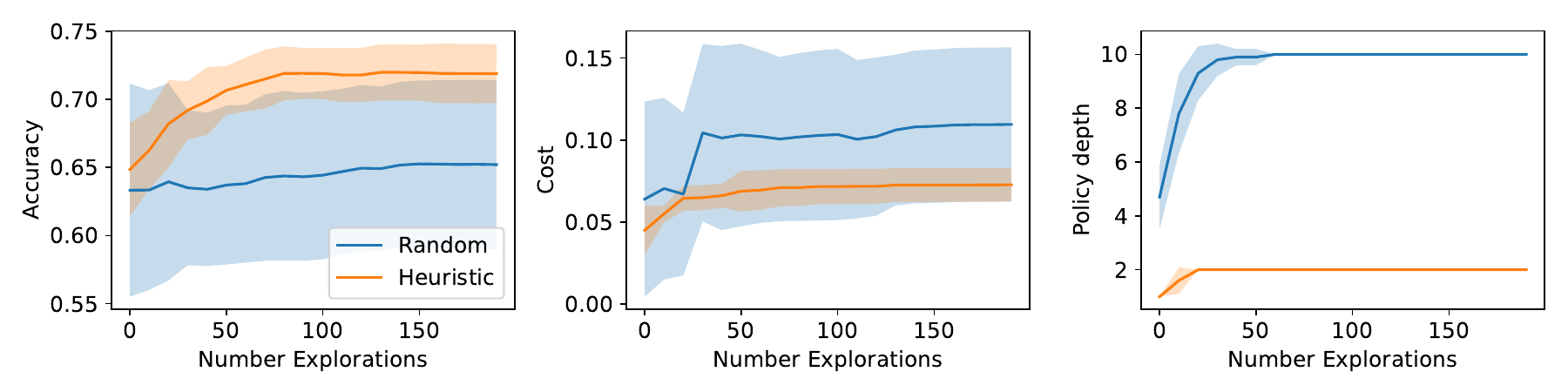}
  \caption{Evolution of the mean return accuracy  (left), mean total cost of all acquired features (center) and policy depth (right) for two exploration strategies: random (blue) and  heuristic (orange), on the "Spam" dataset. The upper row depicts an example with a low acquisition cost per modality ($0.01$ for each modality, while the cost of a misclassification is set to a unit), while the lower corresponds to a greater acquisition cost ($0.05$).
  }
  \label{fig:differentTraining}
\end{figure*}

We tested these  strategies for sequentially generating the policy on the "Spam" dataset. This was repeated for two different feature costs: $10^{-2}$ (low cost), and $5\cdot10^{-2}$ (high cost), reflecting two balances between accuracy and acquisition cost (the cost of a mis-classification is $1$). At each step, we computed the return obtained in an independent test set (Equation~\ref{eq:reward}) as well as the policy complexity, measured by the maximal depth (i.e., the maximal number of acquired modalities) of the found policy. Results are summarized in Figure~\ref{fig:differentTraining}, where we repeated the process over 10 different splits of the training-testing sets for increased statistical stability.

We can observe that random exploration quickly increases the policy depth, while the heuristic explores more nodes from those closest to the root before advancing to the next level, as the maximal policy depth increases slower (Figure \ref{fig:differentTraining}, left). This can be explained as the heuristic overestimates the real value, and does not account for uncertainty in nodes far away from the current policy. This makes that our search strategy promotes exploration over exploitation. When acquisition costs are low and therefore acquiring many modalities is not penalised, a random heuristic outperforms the heuristic strategy. This is because the random exploration rapidly increases the depth as the number of nodes for a given level $k$ is the number of possible combinations of $k$ features, meaning that the number of nodes at each level increases until reaching $N_{features}/2$. 

For higher cost, acquiring many modalities is no longer optimal, and the other two strategies outperform the random search. The heuristic (blue), which explores regions that produce an average high value for all instances, outperforms the dummy random selection. This can be explained because the random strategy can reach solutions that have high accuracy, but requires a large number of features and thus has a high cost. This can be seen in Figure \ref{fig:differentTraining}

\subsection{Regularisation}

We evaluated on the "HTA" dataset the effect of our regularisation technique to simplify policies. Our regularisation penalises policies with a large number of hyperstates visited by at least one sample of the population. The regularisation algorithm behaves similarly to pruning, identifying hyperstates that can be removed without much influence to the global return (either because they are visited by few instances, or because instances visiting them can be de-routed to other superstates that grant a similar return). The sequential process is explained in Section~\ref{sec:policyRegularisation}. It consisted in training first a policy without any regularisation, computing the number of visits to each node following that policy, and subsequently introducing regularisation based on the visits. This process was repeated until convergence, which was attained in less than 5 iterations. 

\begin{figure*}
    \centering
    \subfigure[Non regularised policy]
    {
        \includegraphics[width=14cm,trim=0cm 30cm 0cm 30cm,clip]{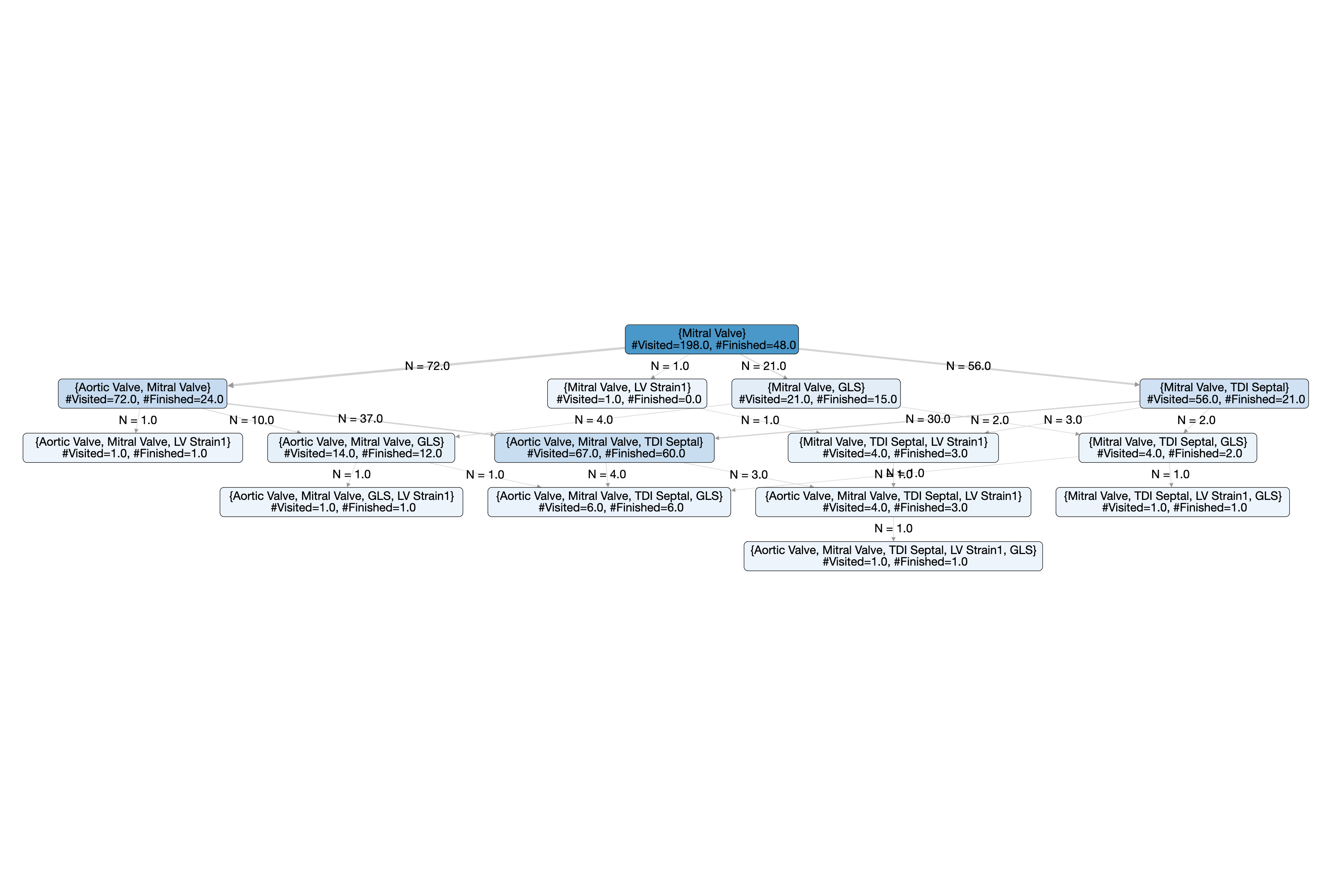}
    }
    \\
    \subfigure[Regularised policy]
    {
        \includegraphics[width=5cm]{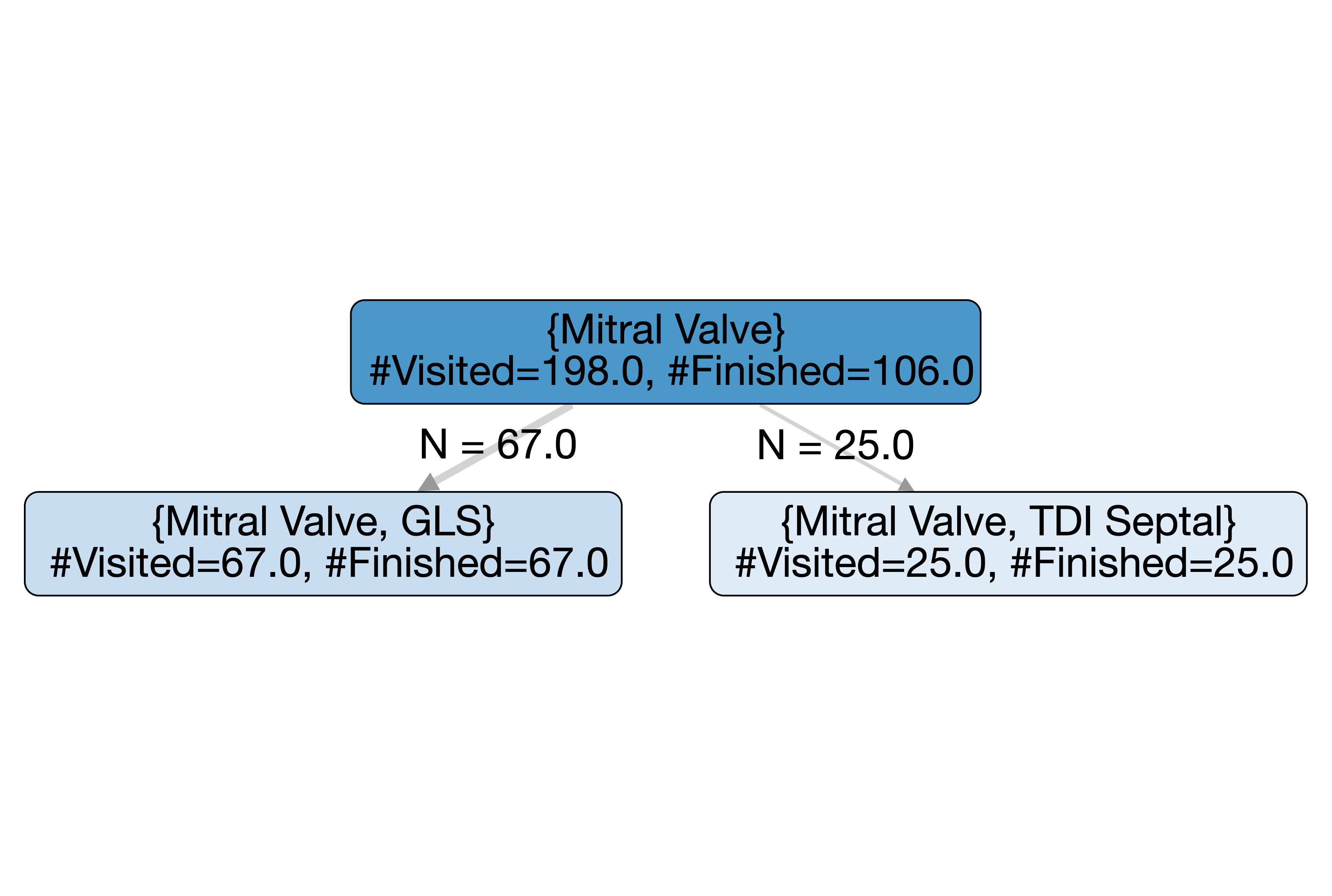}
    }
    \caption{Original (a) and regularised (b) versions of the policy for a given training sample. The graph represents the different modalities acquired by following the resulting policy, and the node color is darker if more instances visited this node. In each node, we also indicated the amount of instances that visited it, and the amount of samples whose final prediction was taken at that node.}
    \label{fig:exp:policies_reg}
\end{figure*}

In Figure \ref{fig:exp:policies_reg}, we can see an example of an unregularised (above) and a regularised policy (below), for the "HTA" dataset. In both  policies, the first modality chosen is the Doppler waveform of the mitral valve, which offers a global view of the diastolic cardiac phase. This modality can detect filling abnormalities, which are well known to be a consequence of an impaired relaxation in hypertensive individuals, as the heart needs to develop more force during ejection to compensate for the increased arterial resistance. As Doppler is a low-cost and readily available modality, it makes sense to be a first choice to try to diagnose. For cases with still an uncertain diagnosis, more modalities were acquired according to the estimated policy. Since the recommended acquisition depends on the patient specificities, this leads to a high number of feature combinations. After regularisation, most of these nodes are removed, and the policy is compacted to only 3 nodes and 2 levels, allowing easier visualisation. 

\begin{figure}
  \includegraphics[width = \columnwidth]{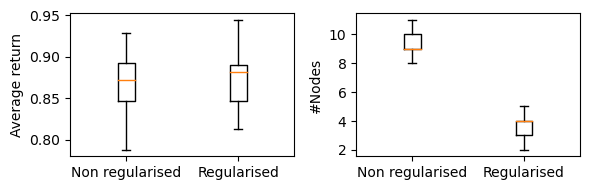}
  \caption{Average return (accuracy minus cost, left) and number of nodes (right). We can see that both returns are comparable, while regularisition significantly decreases the number visited nodes.}
  \label{fig:reg}
\end{figure}

Figure \ref{fig:reg} shows quantitative results on the return in the regularised and unregularised policies, as well as the number of different visited hyperstates. We can see that regularised policies decrease the number of visited superstates, and regularisation does not reduce the total reward (based on the accuracy and cost). 

A challenge in this approach is that the amount of regularisation depends on the hyperparameters $\alpha$ and $q$, that modulate the penalisation the penalisation of using different nodes that have to be selected.

\section{Discussion}

We 
presented 
an original \ac{RL} method for instance-specific feature selection, and demonstrated its relevance on several medical and non-medical datasets (including a dataset with high-dimensional temporal signals). Also, some of these datasets quantified the costs associated to the different modalities acquired or the measurements performed. We proposed a heuristic to reduce the exponentially large search space and specifically evaluated different choices of this heuristic. Furthermore, we proposed and tested a post-processing regularisation technique for pruning the resulting policies, making them more simple (in terms of total amount of different feature combinations) and compact.

\subsection{Comparison to the state-of-the-art}

Our algorithm performed better than state-of-the-art methods in datasets where some of the features are very good predictors, but are expensive to acquire and need to be used scarcely. This was the case for the "Heart" dataset (for which we used the real costs associated to each variable). Several variables correspond to the result of exercise testing, which requires a complex setup and is therefore expensive, but are very informative of the disease. Our algorithm was able to accurately determine individuals that require this costly test, from other individuals that can be accurately diagnosed using cheaper acquisitions.

When only a hard constraint on the number of modalities was used, our algorithm showed similar performance, albeit slightly lower compared to Wang's clustering (which is based on preselecting some possible combinations of variables, common to all individuals). This is because our algorithm is able to choose a different number of modalities for each individual for minimising the average cost. However, when the limitation is just the maximal amount of modalities, the capacity of our method to use less modalities is not needed since the optimal solution will be to acquire always the maximum number of modalities. In datasets with a higher number of variables such as "Spam", our algorithm outperforms Wang's clustering, as their approach limits the amount of different features acquired. 

Our algorithm outperforms EDDI, especially in datasets with multidimensional features. This is because EDDI relies on data imputation, which is not accurate for high dimensional data, such as in the "HTA" dataset. Another influencing factor is that the open implementation available uses a regression loss instead of a classification loss, which can lower the performance of the method.

\subsection{Relationship to the A* algorithm}

Our heuristic shares similarities with the A* algorithm described in \cite{Hart1968APaths}, which finds minimal distance paths in a graph. In A*, there is a set of explored nodes, for which the minimal distance to the origin is computed (for paths containing only explored nodes). There is also a set of "open sets", neighbours of the explored nodes, for which a heuristic is used to estimate the distance to the end-node. At each step, a new node from the open set is selected, based on both the distance to the initial node and the heuristic estimation to the end-node, and added to the explored nodes. This requires to potentially update the distances to the origin of all the explored nodes, so that paths including the new node are also considered; and the unexplored neighbours of the new node are added to the open set, thus their heuristic value needs to be computed at these new neighbours.

In our case, the procedure is similar. We have a set of explored superstates (nodes), whose optimal policy using only those superstates is computed, and an open set of their neighbouring nodes. For the latter, a heuristic based on sampling estimates for each instance the highest reward available from that node. Then, the algorithm selects the open node that offers the highest improvement, based on the heuristic, and updates the policy with the addition of the new superstate, as well as adding new neighbours to the open set.

A problem of our heuristic is that it is a optimistic estimator, specially in the superstates closer to the root. It assumes that there is an oracle able to decide for each instance the best set of features that must be acquired to obtain the highest reward. In practice, there will be some error and instances that will end in sub-optimal combinations, decreasing the achieved reward. This makes that the estimated value will decrease for deeper nodes, which makes that the algorithm will prefer exploring the shallow parts of the graph, instead of exploiting found directions. A possible solution to this would be to include a correction term to account for this effect, similar to the Upper Confidence Bounds in Monte Carlo tree search \cite{Kaufmann2012OnProblems}. However, this term needs to be derived for this specific problem.

\subsection{Regularisation}

A second challenge is the complexity of the policies, i.e. the number of different superstates. Sometimes, when there are features encoding approximately the same information, the optimal policy chooses one or the other almost arbitrarily, increasing the complexity of the policy without producing a significant improvement of the reward. We proposed a regularisation term to promote simpler policies. A challenge in the formulation is that in the \ac{MDP} formulation, reward has to be defined at an instance level, whereas we want to minimise the occupancy  at specific superstates (i.e., the probability over the full population that a certain state is visited through an episode). This requires an iterative alternate optimisation, in which we first estimate the occupancy of each node following a fixed policy, and then add a virtual reward to update the policy based on the previous "fixed" occupancy. This can be derived based on the \ac{REPS} formalism \cite{Peters2010RelativeSearch}. A challenge is that this approach consists of the maximisation of a convex energy, for which only convergence towards a local maxima is guaranteed: in particular, once a superstate reaches 0 occupancy, it is discarded and will no longer be possible to be visited. Therefore, our algorithm has similarities to post-processing pruning algorithms in decision trees \cite{Breslow1997SimplifyingSurvey}.

\section{Conclusion}

We proposed a RL-based method to find instance-specific optimal combinations of features, accounting for both accuracy and feature acquisition cost, to work with a higher number of variables ($\sim$ 10 to 50); and a post-training regularisation procedure to reduce the total number of different feature combinations used in the population. We evaluated our algorithm in different datasets, demonstrating better performance compared to state-of-the-art methods.

A sampling-based heuristic was used to obtain the next superstate to open, based on the one which will grant maximal value. However, this can lead to an excessive exploration of the super-states, as the maximal possible value is overestimated. This heuristic is simple, and can be improved by adding, for instance, a penalisation term to account that the nodes far away from the current explored subgraph have more uncertainty. Adding such a term will promote exploitation \cite{Coquelin2007BanditSearch}.

Finally, we developed a way of compacting policies by reducing the number of different nodes visited the population, allowing an easier inspection and validation of the policy without losing accuracy/cost.


\appendix
\section{Derivation of the sparse-promoting regularisation term}
\label{sec:AppendixA}

We use the smoothed version of the regularisation term, for a fixed value of $q$, and  we use the \ac{REPS} formulation \cite{Peters2010RelativeSearch}. In this equivalent formulation of the \ac{MDP}, instead of optimising over the decision functions, the optimisation is done in dual variables, which correspond to the occupancy functions $\mu$. We will add our regularisation term $H^q$ to the objective:
\begin{equation}
\begin{array}{ll@{}ll}
\text{maximize}  & \displaystyle \langle r, \mu \rangle + \alpha H^q(\mu), &\\
\text{subject to}&  E^T\mu = \gamma P^T \mu + (1 - \mu) \nu_0, \\
\end{array}
\end{equation}

where the constraint equation assures that the transitions are satisfied ($E$ and $P$ are the transition functions) and $\nu_0$ is the initial distribution.

This problem is typically solved with a primal-dual proximal operator algorithm. After linearising the regularisation term at $\mu_k$,  ($H^q(\mu) \approx H^q(\mu_k) + \langle \nabla H^q(\mu_k), \mu - \mu_k \rangle $) we obtain:
\begin{equation*}
\begin{array}{ll@{}ll}
\text{maximize}  & \displaystyle \langle r, \mu \rangle + \alpha (H^q(\mu_k) + \langle \nabla H^q(\mu_k), \mu - \mu_k \rangle) &\\
& + \mathcal{O}(\|\mu - \mu_k\|^2),&\\  & \\
\text{subject to}&  E^T\mu = \gamma P^T \mu + (1 - \mu) \nu_0. \\
\end{array}
\end{equation*}

For small enough policy updates, the quadratic term can be neglected. Moreover, after reorganising the terms, and removing those that only depend on $\mu_k$, since they do not contribute to the maximisation result, the following problem is obtained:
\begin{equation*}
\begin{array}{ll@{}ll}
\text{maximize}  & \displaystyle \langle r + \alpha \nabla H^q(\mu_k), \mu \rangle,  & \\
\text{subject to}&  E^T\mu = \gamma P^T \mu + (1 - \mu) \nu_0. \\
\end{array}
\end{equation*}

This leads to a formulation in which, at each step, the gradient of the regularisation term at the current state-occupancy measure is added to the rewards as a "virtual" reward, which equals our Equation \ref{eq:virtualReward}.

\section*{Acknowledgment}

The authors acknowledge the partial support from the LABEX PRIMES of Université de Lyon (ANR-11-LABX-0063), the French ANR (MIC-MAC project, ANR-19-CE45-0005), the Fédération Francaise de Cardiologie (MI-MIX project, Allocation René Foudon), the Institut Universitaire de France, La Marato TV3 (202415 30 31) the European Union-NextGenerationEU, Spanish Ministry of Universities and Recovery, Transformation and Resilience Plan, through a call from Pompeu Fabra University (Barcelona), TAILOR (EU H2020 
\#952215), AGAUR SGR and Spanish grants PID2019-108141GB-I00, PID2023-149959OA-I00, RYC2022-035960-I,  and the Maria de Maeztu Units of Excellence Programme (CEX2021-001195-M).




\bibliographystyle{plain}
\bibliography{references}

\end{document}